\documentclass[runningheads]{llncs}
\usepackage{graphicx}
\usepackage{hyperref}
\usepackage{booktabs}
\usepackage{multirow}
\usepackage[table, dvipsnames]{xcolor}
\usepackage{hhline}
\usepackage{subcaption}

\begin{document}

\title{The Socface Project: Large-Scale Collection, Processing, and Analysis of a Century of French Censuses}
\titlerunning{The Socface Project}

\author{Mélodie Boillet\inst{1}\orcidID{0000-0002-0618-7852} \and
Solène Tarride\inst{1}\orcidID{0000-0001-6174-9865} \and
Manon Blanco\inst{1} \and
Valentin Rigal\inst{1} \and
Yoann Schneider\inst{1} \and
Bastien Abadie\inst{1} \and
Lionel Kesztenbaum\inst{2} \and
Christopher Kermorvant\inst{1}\orcidID{0000-0002-7508-4080}
}
\authorrunning{M. Boillet et al.}

\institute{TEKLIA, Paris, France \\
\email{boillet@teklia.com} \\
\and Institut National d'Etudes Démographiques (INED) and Paris School of Economics (PSE), France\\
\email{lionel.kesztenbaum@ined.fr}
}

\maketitle              
\setcounter{footnote}{0} 

%
\begin{abstract}

This paper presents a complete processing workflow for extracting information from French census lists from 1836 to 1936. These lists contain information about individuals living in France and their households. We aim at extracting all the information contained in these tables using automatic handwritten table recognition. At the end of the Socface project, in which our work is taking place, the extracted information will be redistributed to the departmental archives, and the nominative lists will be freely available to the public, allowing anyone to browse hundreds of millions of records. The extracted data will be used by demographers to analyze social change over time, significantly improving our understanding of French economic and social structures. For this project, we developed a complete processing workflow: large-scale data collection from French departmental archives, collaborative annotation of documents, training of handwritten table text and structure recognition models, and mass processing of millions of images.

We present the tools we have developed to easily collect and process millions of pages. We also show that it is possible to process such a wide variety of tables with a single table recognition model that uses the image of the entire page to recognize information about individuals, categorize them and automatically group them into households. The entire process has been successfully used to process the documents of a departmental archive, representing more than 450,000 images.


\end{abstract}

%
\section{The Socface project}
\label{sec-introduction}

The Socface project\footnote{\url{https://socface.site.ined.fr/}} involves archivists, demographers, and computer scientists working together to analyze French census documents and extract information on a very large scale. Its objective is to gather and process all the handwritten nominal census lists from 1836 to 1936 using automatic handwriting recognition. Produced every five year, these lists are organized spatially (municipality; wards, hamlets, or streets; houses; households) and summarize the information from the census, listing each individual with some of his or her characteristics, e.g., name, year of birth, or occupation. The project aims at taking advantage of this archival material to produce a database of all individuals who lived in France between 1836 and 1936, which will be used to analyze social change over the course of 100 years. An important impact of Socface will be public access to the nominal lists: they will be made freely available, allowing anyone to browse hundreds of millions of records.

\begin{figure}[ht]
    \centering
    \begin{subfigure}[b]{0.31\textwidth}
    \centering
         \includegraphics[height=5.5cm]{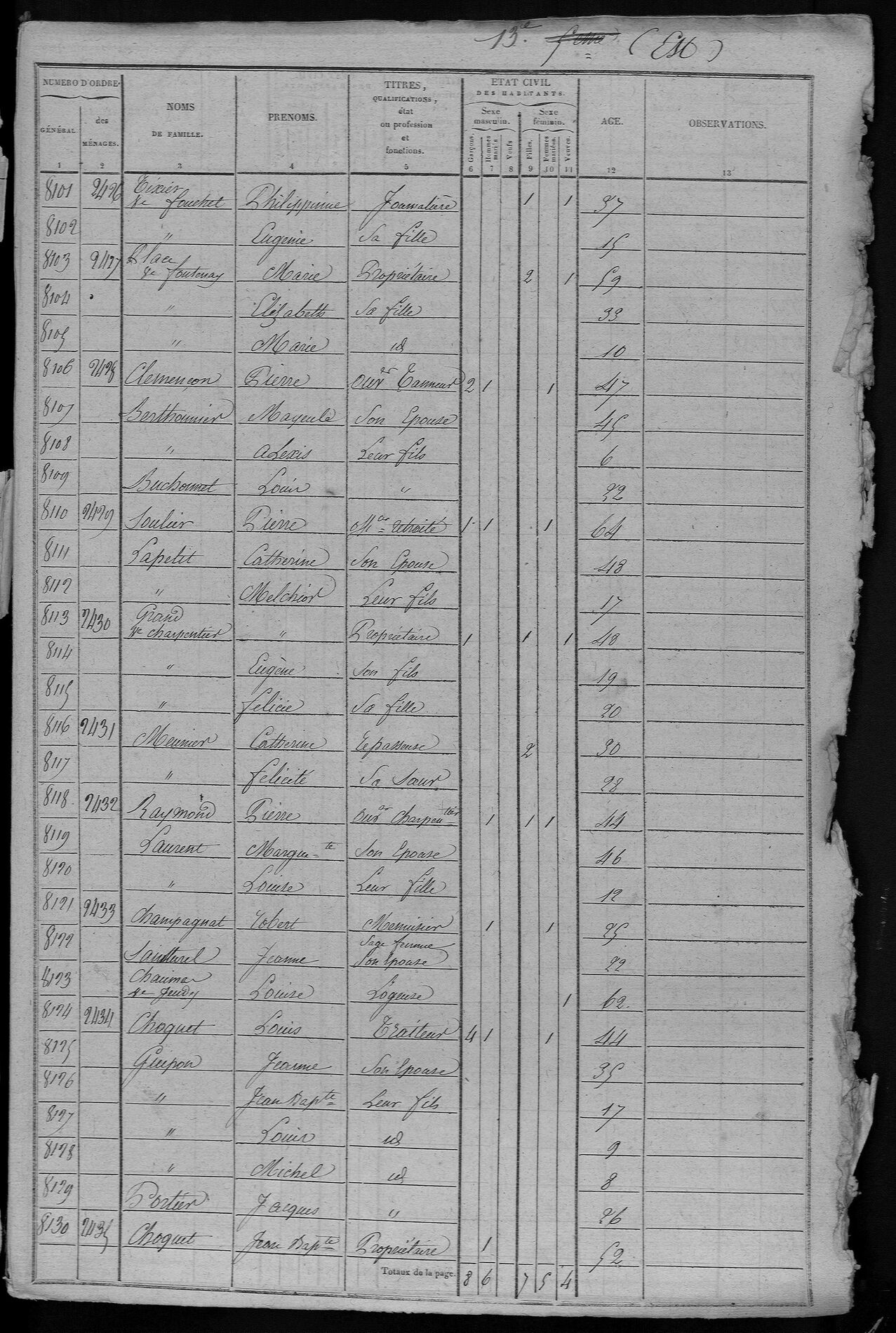}
         \caption{1836}
         \label{fig:moulins_1836}
    \end{subfigure}
    \begin{subfigure}[b]{0.31\textwidth}
    \centering
         \includegraphics[height=5.5cm]{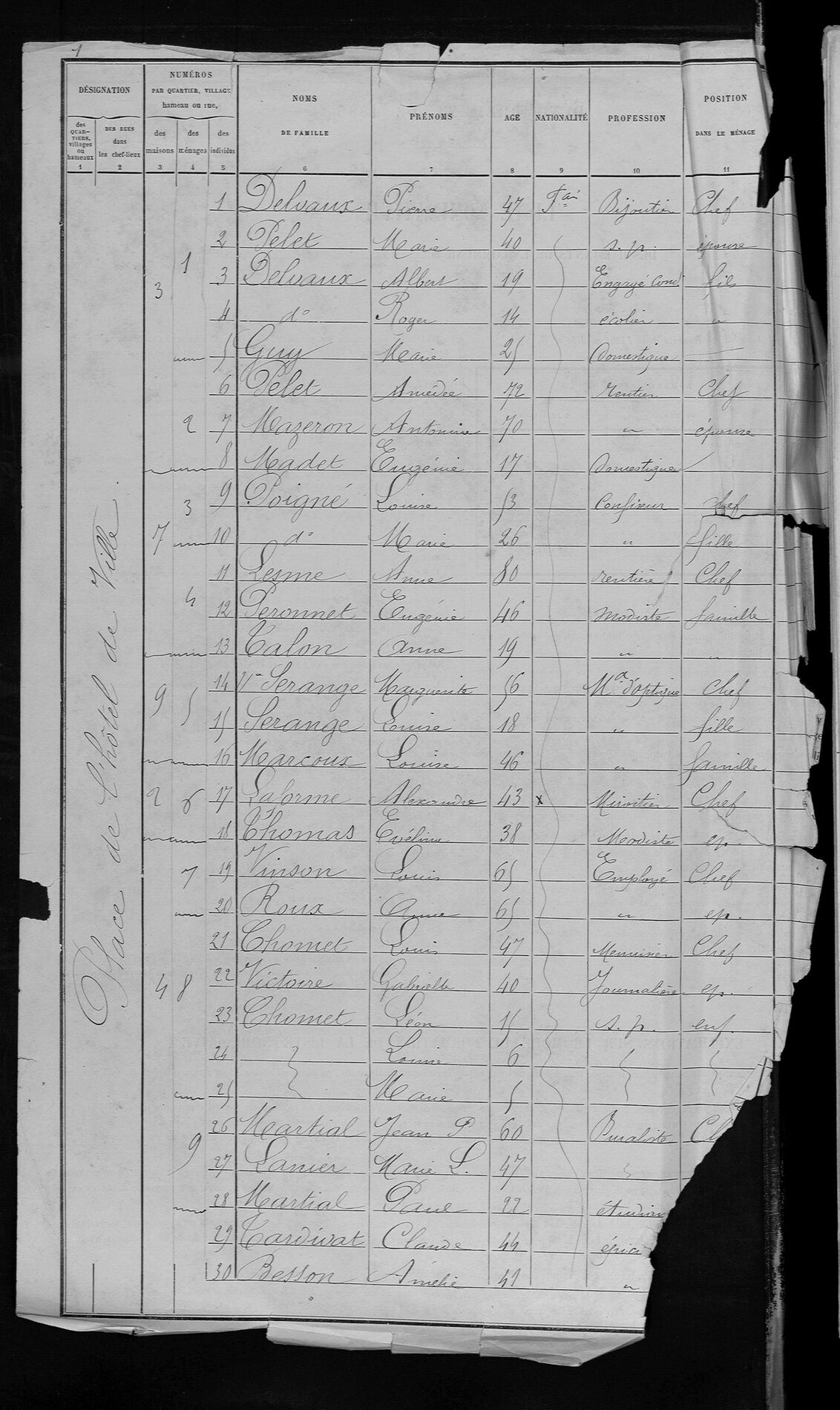}
         \caption{1886}
         \label{fig:moulins_1886}
    \end{subfigure}
    \begin{subfigure}[b]{0.31\textwidth}
    \centering
         \includegraphics[height=5.5cm]{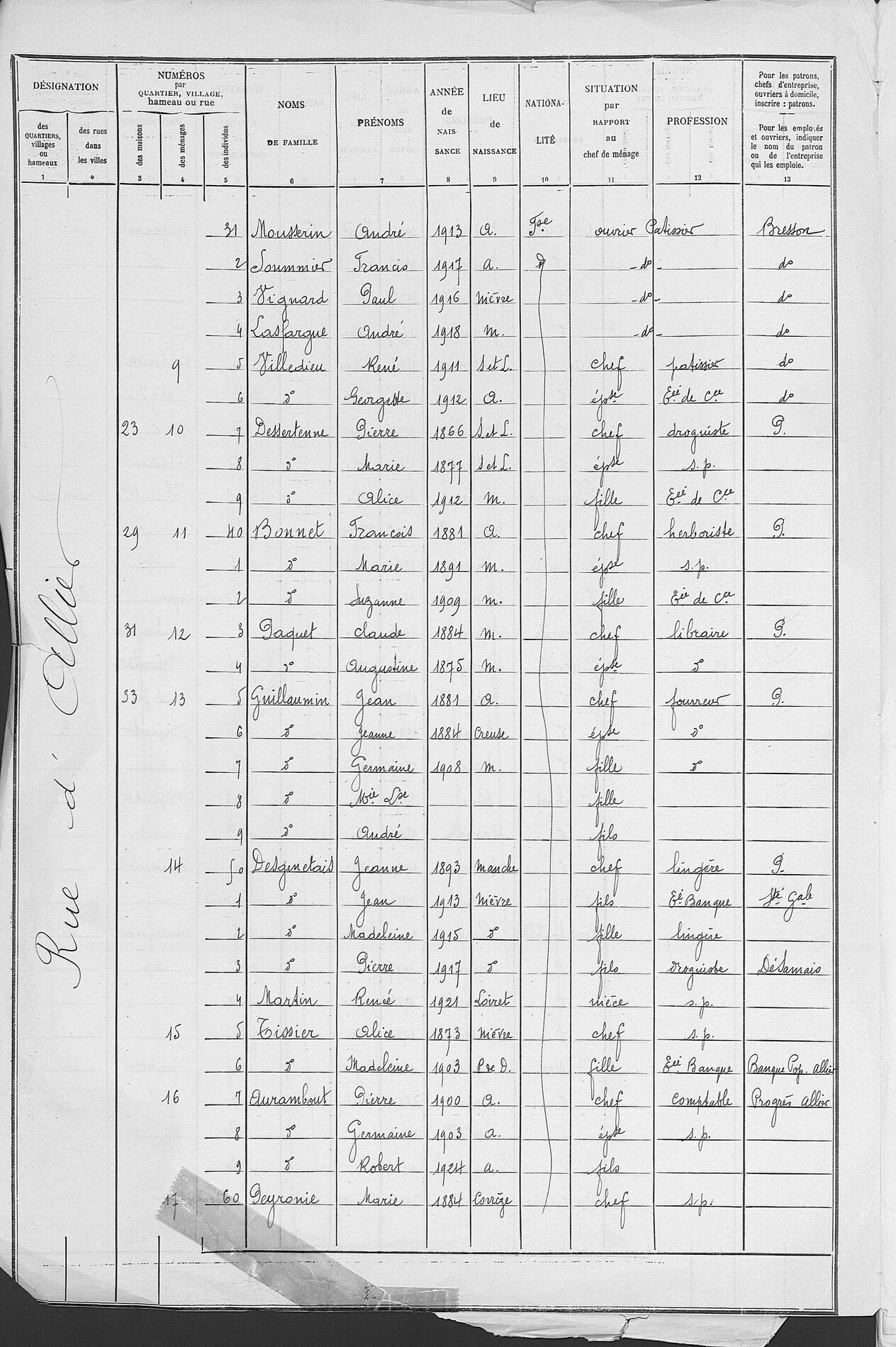}
         \caption{1936}
         \label{fig:moulins_1936}
    \end{subfigure}
    \caption{First page of nominal lists for the commune of Moulins (department of Allier) for three census years. The quality of the pages varies greatly from one year to the next. In addition, the table template evolved over the years: in 1881, civil status was replaced by the column marking the position in the household. In 1906, age is replaced by year of birth, as can be seen on the 1936 example.}
    \label{fig:example_images}
\end{figure}

As depicted on Figure \ref{fig:example_images}, the data are presented in tabular form. A major challenge in this project and in processing these documents is that the tabular formats have evolved over the 100 years studied. As can be seen in the figure, the columns changed (age vs. year of birth), so did their order on the page. In addition, the quality of preservation varies from year to year and from archival deposit to archival deposit. The very large number of writers makes the task even more complex. 

The decentralized nature of the source material has been a significant hurdle for prior attempts at a project of this scale. The images required for the Socface project are dispersed across 100 local archive services throughout France, rather than being housed in a single repository. The project is faced with a dual layer of variability due to the dispersion of documents, which requires not only the preliminary collection of images, but also dealing with the diversity of the documents themselves and the differing organizational systems and metadata standards employed by each archival service. The collection and systematic analysis of the data is made difficult by its complexity. 

The Socface project faces a significant challenge in processing a vast number of documents, with an estimated count of 30 million images. To address this challenge, access to public supercomputing resources is necessary. However, High-Performance Computing (HPC) architectures are not inherently designed to manage such extensive input and output flows. Tailored development efforts are necessary to ensure that images can be efficiently processed by available computing resources, particularly GPUs, and to seamlessly integrate the results into a document management system. This highlights the need for innovative solutions to bridge the gap between traditional HPC capabilities and the demands of large-scale data analysis projects. \\


In this paper, we describe the methodologies and technological advancements developed in the Socface project, highlighting our contributions to document recognition and historical data analysis. Our work presents a comprehensive approach to processing and analyzing historical census documents on an unprecedented scale. The key contributions of this paper are:
 
\paragraph{Data collection and normalization:}
 We present a reliable method for collecting, identifying, normalizing, and storing images and metadata from each archival service. This involved developing a standardized protocol for interacting with the various organizational systems found across the 100 local archives, ensuring consistency in the way documents are digitized, classified, and archived. Our approach involves techniques for harmonizing metadata, which facilitates the integration of different data sources into a cohesive dataset.
 
\paragraph{Deep learning model for handwritten table understanding:}
 A central contribution of our work is the design of a unique deep learning model capable of recognizing and structuring the personal information contained within the handwritten lists, despite the considerable diversity of document formats encountered. This model leverages advancements in full page handwriting recognition to accurately interpret a wide range of handwriting styles and extract structured data from documents whose layouts evolve over time. 
\paragraph{ High-Performance Computing (HPC) for document processing:}
 A pivotal advancement made in this project is the extension of Arkindex, an open-source platform for automatic document processing, to facilitate communication with High-Performance Computing (HPC) systems via the SLURM workload manager. This extension grants  the document processing community the ability to leverage the vast processing capacities inherent to HPC environments.

This paper is structured as follows. Section \ref{sec-related-work} provides an overview of the main approaches for information extraction from digitized handwritten tables. Section \ref{sec-data-collection} presents the tool developed during this project to simplify the data collection and normalization from departmental archives. Section \ref{sec-data} presents the census registers and the annotation process. Section \ref{sec-ie-workflow} describes the training data, presents each step of the proposed information extraction workflow, and discusses the results. The final section \ref{sec-full-corpus-processing} describes how we distributed the document processing across a cluster of computers using HPC tools.

%
\section{Related work}
\label{sec-related-work}


Several models are available for detecting tables in document images. However, there are few models that can retrieve both the textual content and structure of tables, especially for historical and handwritten documents. The dominant approach for processing such documents is to first detect the rows of a table and then apply a standard character recognition model at the row level. More recently, models have been proposed to process handwritten tables as a whole by analyzing the image of the entire table. Both approaches will be discussed in the following sections.

\subsubsection{Table row processing}

In the literature, most analysis focus on 2-step pipelines. First, the table rows are extracted using standard text line detection models. These are usually Fully Convolutional Networks (FCN) \cite{gruning2018,boillet2020}, Region-based CNNs (R-CNN) \cite{ren2015,he2015} or, more recently, Transformer-based \cite{vaswani2017,biswas2022} models. Once the rows have been extracted, standard text recognition using an HTR model is applied, and the columns are often recreated in a post-processing step.

In their work on the POPP dataset, Constum \textit{et al.} \cite{constum2022} addressed the problem using a standard line detection model \cite{oliveira2018} followed by a line-level text recognition model \cite{coquenet2023-1}. As their tables followed the same template from image to image, there was no need to segment the tables into columns, as the information was always presented in the same order. To correctly categorize the retrieved information, they added a \texttt{/} symbol in the ground truth to separate the information from the different columns.

In their study, Tarride \textit{et al.} \cite{tarride2023-1} made this method a little more generic by predicting both the text and the category of information recognized. This makes it possible to apply the same model to multiple table templates. To achieve this, the authors transformed the information in the ground truth by adding a token before the start of the text in each column, representing its category. This allowed them to avoid using the \texttt{/} symbol, which was no longer useful. The trained model performs very similarly to the model trained by \cite{constum2022}, but it is much more general and predicts more information as it categorizes the detected information.

The TableTransformer model \cite{smock2022}  goes one step further by extracting both tables and their structure from PDF document images. This means that it can extract not only the rows of tables, but also their columns and cells. This model works very well on printed data and has shown good performance on handwritten tables \cite{bernard2023}. However, as with previous approaches, it cannot recognize the content of the cells directly, so it is necessary to apply a text recognition model afterward.

\subsubsection{Full table processing}

A major disadvantage of processing at the table row level is that, as with conventional text recognition, detection errors have a major impact on the quality of text recognition. Furthermore, if we use a character recognition model that uses context, in particular Transformer models, the context is greatly reduced compared to full page recognition. This is also the case for table processing: when processing at row level only, the context of previous rows is lost, as is the information contained in the table header. Recent advances, particularly in Transformer architectures, make it possible to process and understand the entire page or table without any prior segmentation. 

In \cite{tarride2023-1}, the authors trained a model on the whole table. The model has been trained to extract the textual content of the table in a structured way: each piece of textual information is extracted with its type, which corresponds to the column label. For this, the authors used the DAN model \cite{coquenet2023-2}, which combines a convolutional encoder and a transformer decoder, making it possible to process documents at line level, but also at full page level. Due to the small amount of annotated data and the much more complex task, the model performs less well than the table row level model, but is not affected by the quality of the line-detection model, whose impact on text recognition has not been assessed.


Given the immense size and diversity of the documents in the Socface project, it is impractical to set up a complex workflow with sequential models such as in \cite{tarride2023balsac} or to use several models tailored to specific document templates. We therefore decided to develop a single, comprehensive model capable of processing entire tables. This model is designed to automatically adapt to the variations inherent in documents, ensuring efficient and accurate recognition and structuring of data without the need for prior segmentation or template-specific adjustments. This approach not only streamlines the processing pipeline, but also overcomes the challenge of handling the project's large and varied dataset with a scalable and flexible solution.

\section{Data collection and normalization}
\label{sec-data-collection}

\begin{figure}[t]
    \centering
    \includegraphics[width=0.6\linewidth]{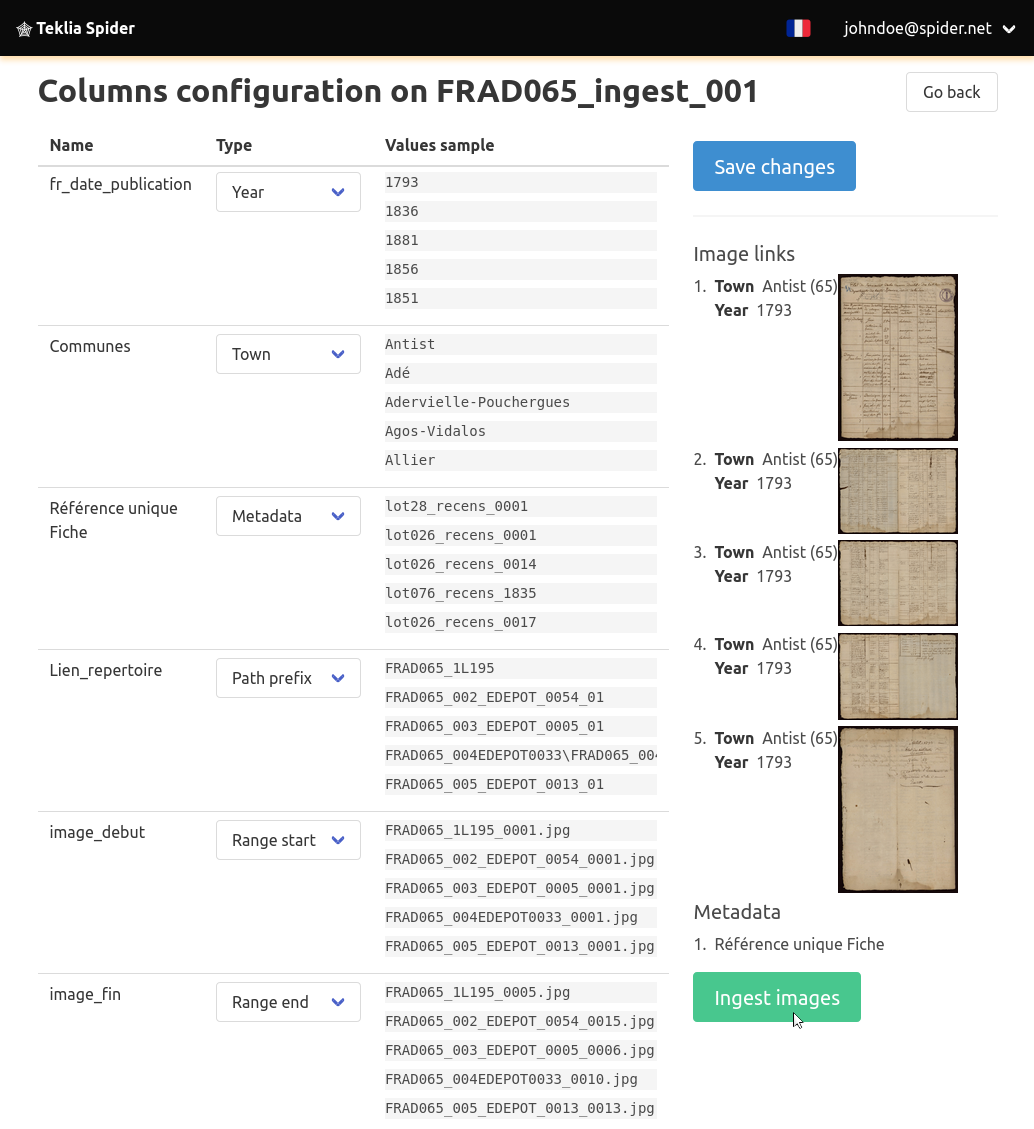}
    \caption{Configuration interface for retrieving and organizing data from the input CSV file. The "Name" column indicates the fields present in the CSV file. The "Type" column indicates how the CSV fields will be used (whether it corresponds to the year or commune, or if the field should be ignored). If the data displayed in the "Values sample" column is correct, the user will see a preview of the retrieved images with their metadata.}
    \label{fig:spider}
\end{figure}

A critical component of the Socface project is the comprehensive collection, normalization, and organization of images and metadata from 94 departmental archives services across metropolitan France. The majority of these services have volunteered to participate in the project by providing access to their archival images and associated metadata. In return for their cooperation, they are offered access to all the data automatically extracted from their documents. These archive services use a variety of systems to store their images, including self-hosted solutions, external hosting and the International Image Interoperability Framework (IIIF), as well as different archive management systems. As a result, images and metadata were presented in a variety of formats and organizational hierarchies, including XML-EAD, CSV, XLSX, XLS and ODT, with no standard naming conventions for cities across the services.

To address the challenge of collecting, organizing and normalizing this diverse dataset, we developed a web-based platform called Socface-Spider. This platform is designed with several key functionalities to facilitate the processing of the collected data:
\begin{itemize}
\item Import metadata from CSV files: In response to the diversity of file formats provided by the archive services, all file formats are first converted to CSV. Following conversion, metadata files are imported into Socface-Spider.
\item Support for specific CSV formats: Given the variation in the structure of CSV files across different archive services, Socface-Spider includes a feature that allows users to manually select the columns containing the necessary metadata. This selection is facilitated by a user interface, shown in Figure \ref{fig:spider}, designed to accommodate the specificities of each CSV format. This process ensures that essential data such as the year, city name, archival ID, and image path are accurately identified and normalized for consistency.  
\item Fuzzy identification of place names: Given the lack of standardized naming of cities across services, the platform uses fuzzy matching techniques to identify city names within the Cassini index \cite{motte2007}. This index catalogues all official names of communes in France since 1793, facilitating accurate matching of data to specific locations.
\item Image integrity checks via IIIF: The platform verifies the presence and integrity of images on the storage server via IIIF access, ensuring that digital artifacts are complete and uncorrupted before further processing.
\item Export and organization of validated data: After validation, the platform exports the data to Arkindex, where the images are organized in a standardized manner by census year, municipality, and register. All the metadata collected is linked to the corresponding census registers, creating a structured and accessible dataset.
\end{itemize}

At the current stage of the project, Socface-Spider has proven its effectiveness and versatility by being used in more than 50 projects, successfully validating and organizing more than 9 million images and their metadata according to the specific requirements of each project.

\section{Document organization and content}
\label{sec-data}

\subsection{Description of census registers}
\label{sec-data-description}

The census registers provide a unique window on the demographic fabric of France from the mid-nineteenth to the mid-twentieth century. These nominative lists were systematically compiled every five years from 1836 onwards. Exceptions to this five-year rhythm were due to historical contingencies: the census of 1871 was postponed to the following year due to the occupation of parts of the territory by the Prussian army, and those planned for 1916 and 1941 were cancelled due to war conditions. The censuses were carried out within the municipal framework, systematically listing the inhabitants by household. This organization gave priority to the head of the household, followed by his or her spouse, children, other relatives living in the household, and then any servants or apprentices, among others.

\begin{figure}[ht]
    \centering
    \begin{subfigure}[b]{0.24\textwidth}
    \centering
         \includegraphics[height=4.2cm]{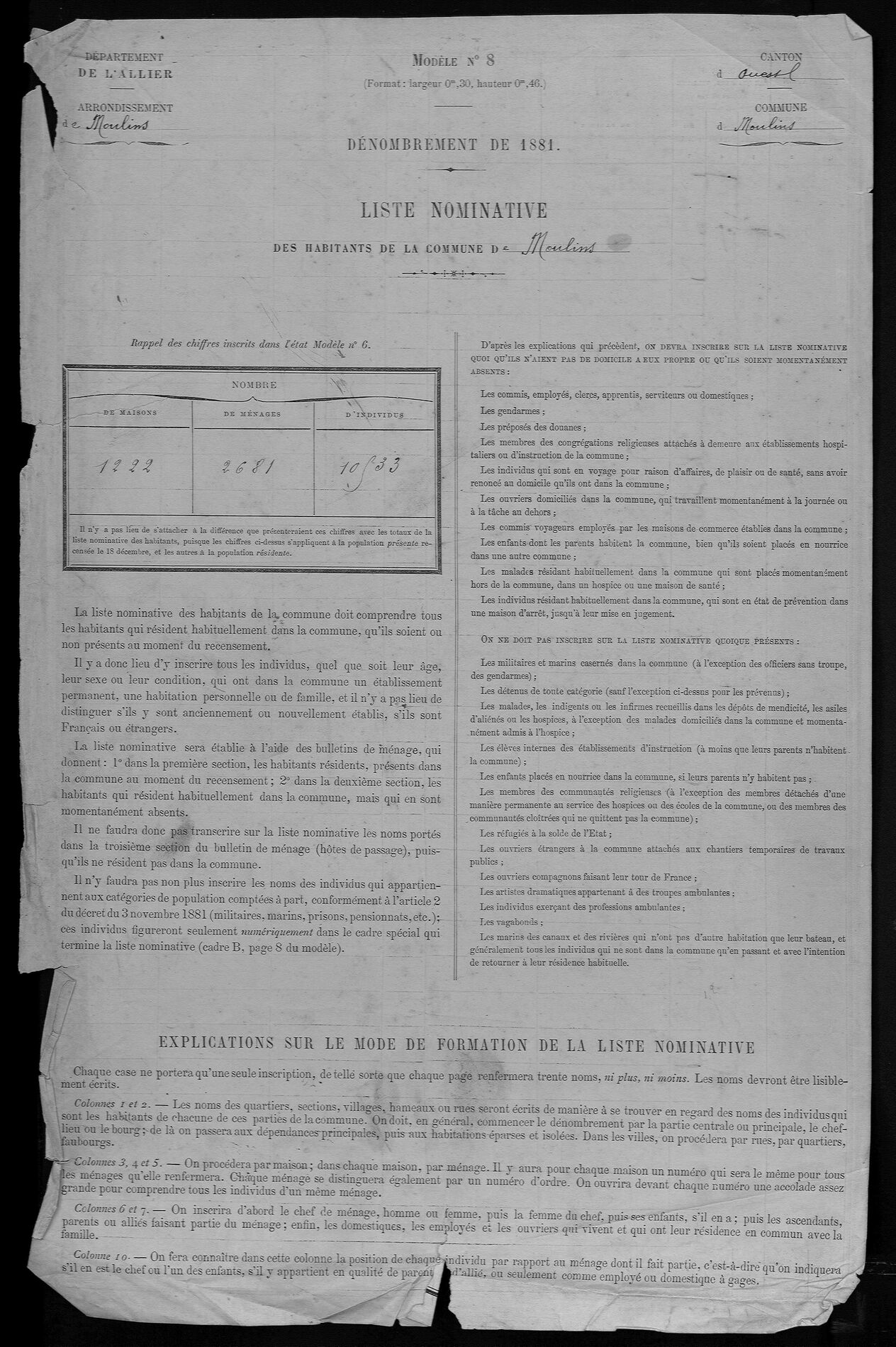}
         \caption{\textsc{Front page}}
         \label{fig:front_page}
    \end{subfigure}
    \begin{subfigure}[b]{0.24\textwidth}
    \centering
         \includegraphics[height=4.2cm]{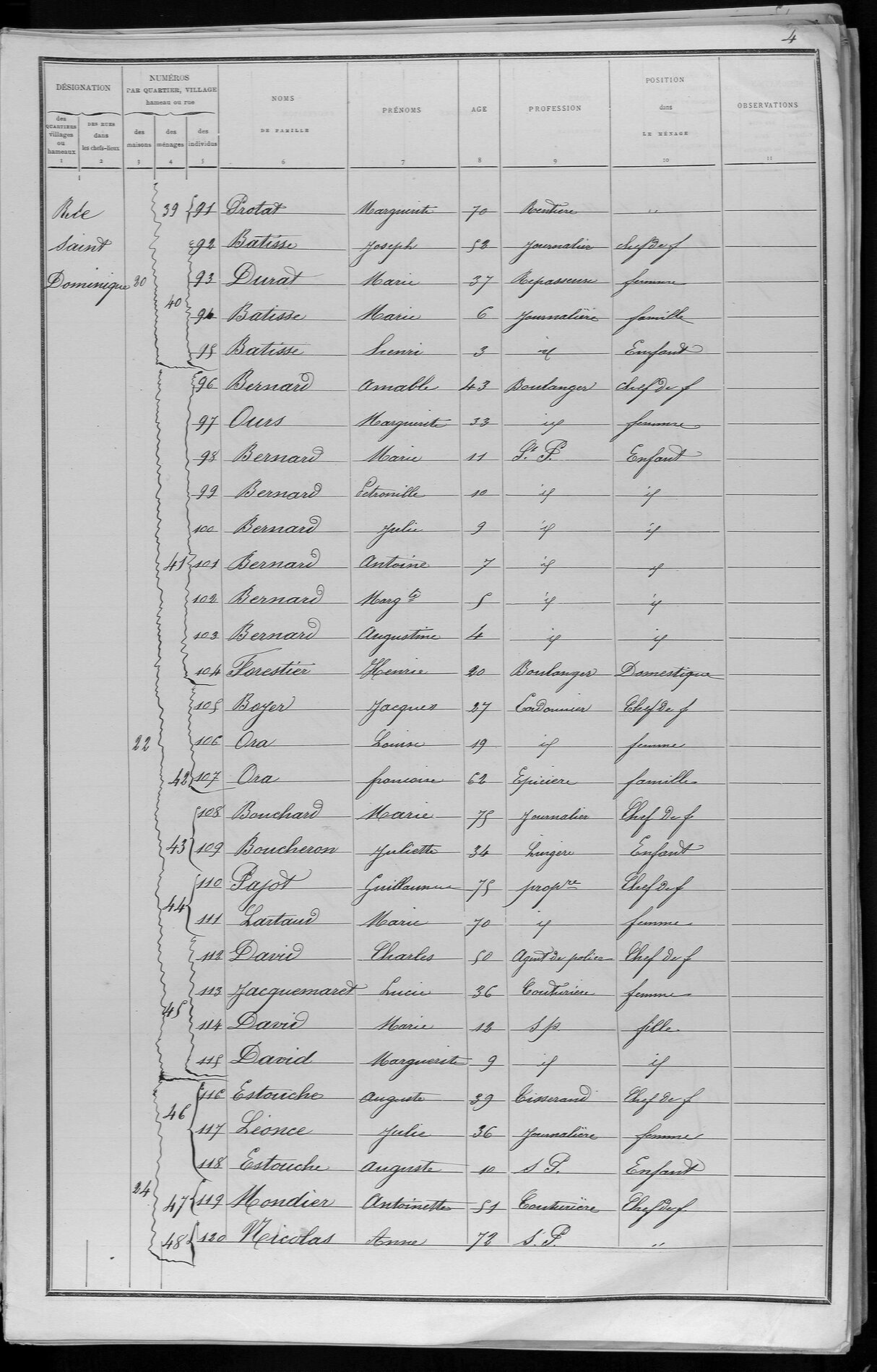}
         \caption{\textsc{List page}}
         \label{fig:list_page}
    \end{subfigure}
    \begin{subfigure}[b]{0.24\textwidth}
    \centering
         \includegraphics[height=4.2cm]{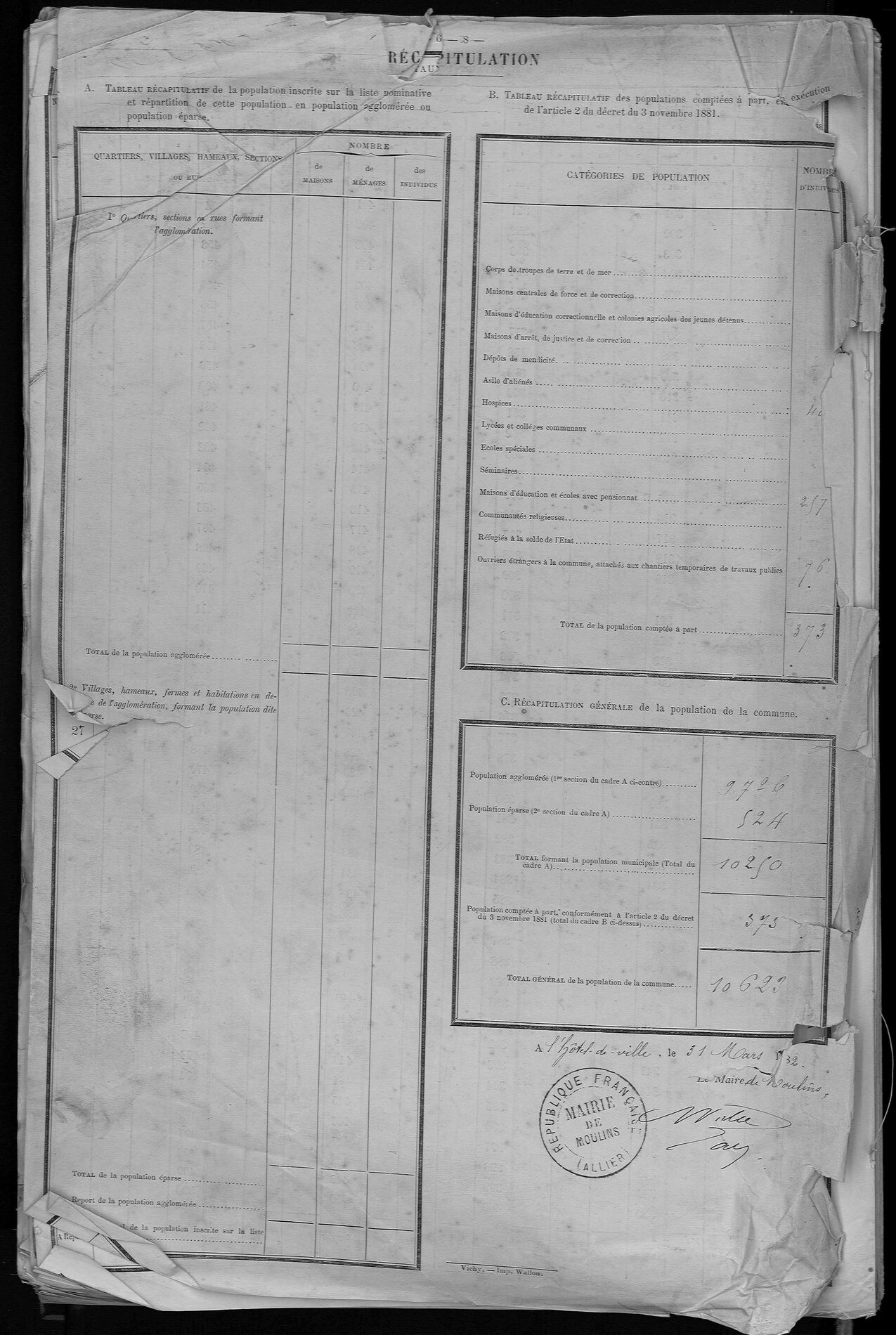}
         \caption{\textsc{Recap page}}
         \label{fig:recap_page}
    \end{subfigure} 
    \begin{subfigure}[b]{0.24\textwidth}
    \centering
         \includegraphics[height=4.2cm]{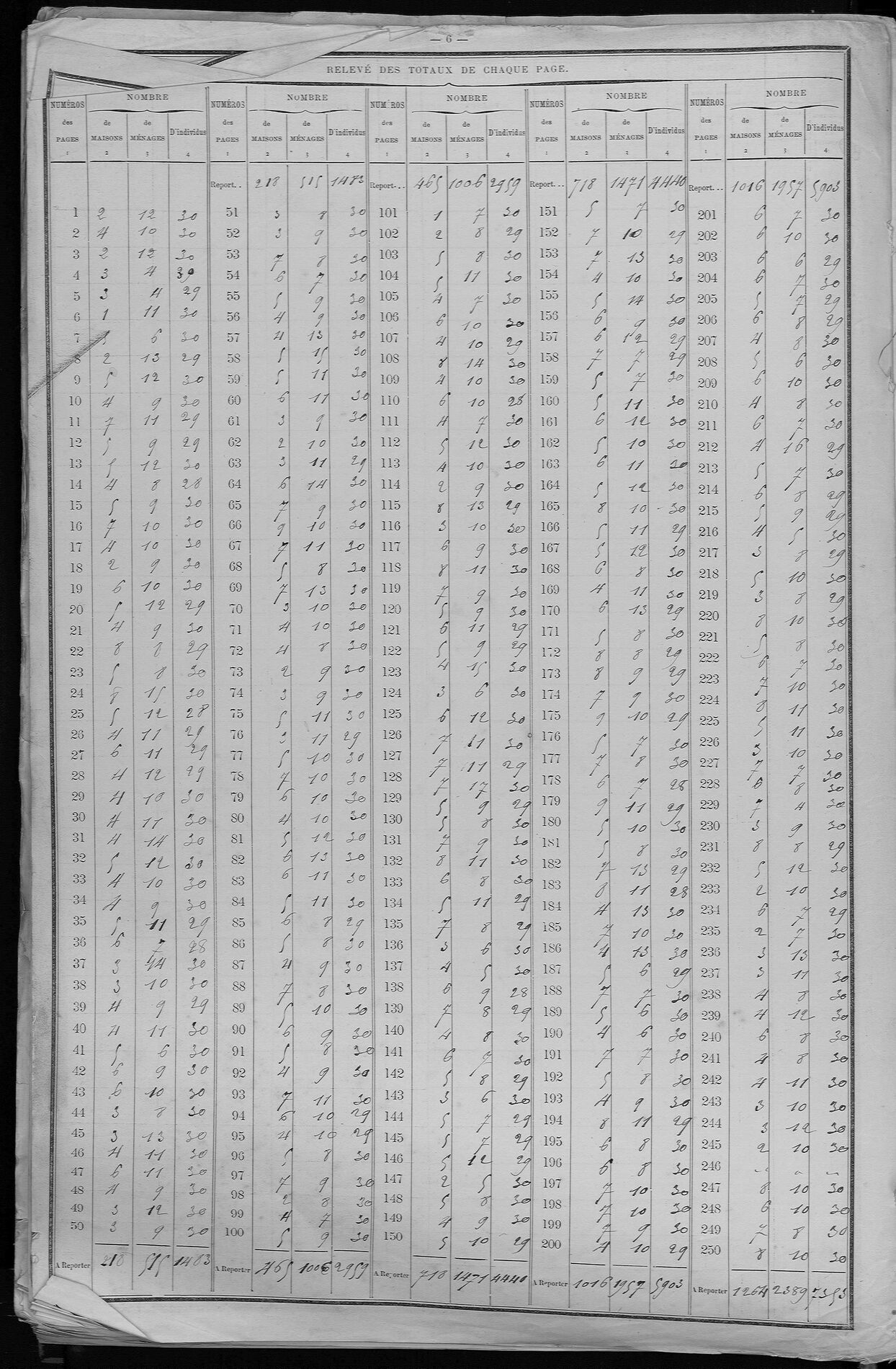}
         \caption{\textsc{Totals page}}
         \label{fig:totals_page}
    \end{subfigure}
    \caption{Example of digitized pages from the census of the commune of Moulins (department of Allier) in 1881.}
    \label{fig:moulins_example_images}
\end{figure}

Over time, the content of these communal nominative lists evolved and typically included first and last names, ages or dates of birth, family positions, occupations, nationalities, and occasionally precise addresses. The images obtained from the archival services are systematically organized into registers, each corresponding to a specific census date and commune. The images are mainly scans of double pages and, for certain years and departments, single pages.  These include not only the nominative lists, but also title pages, summaries, totals and even blank pages as presented in Figure \ref{fig:moulins_example_images}, with most images in black and white, scanned either from the originals or from microfilm, although a few are in color.

At present, our project is concentrating exclusively on the pages containing individual information organized by household. These lists are usually 30 lines long, although variations from 29 to 36 lines have been observed.

The layout of these documents generally begins with columns for street, house and household information, followed by details of individuals such as surname, first name, age (or year of birth) and occupation. At the current stage of the project, we are focusing on the recognition and analysis of the individual information contained in these lists.

\subsection{Ground-truth generation}
\label{sec-callico}

Generating ground-truth data is a fundamental step in training deep learning models for automatically extracting individual information from historical census lists. Given the wide variation in documents - including differences in time, format, and scanning conditions - it is imperative to collect and annotate a representative sample that captures the full spectrum of document diversity. 

\begin{figure}[ht]
    \centering
    \begin{subfigure}[b]{0.48\textwidth}
         \centering
         \includegraphics[width=\linewidth]{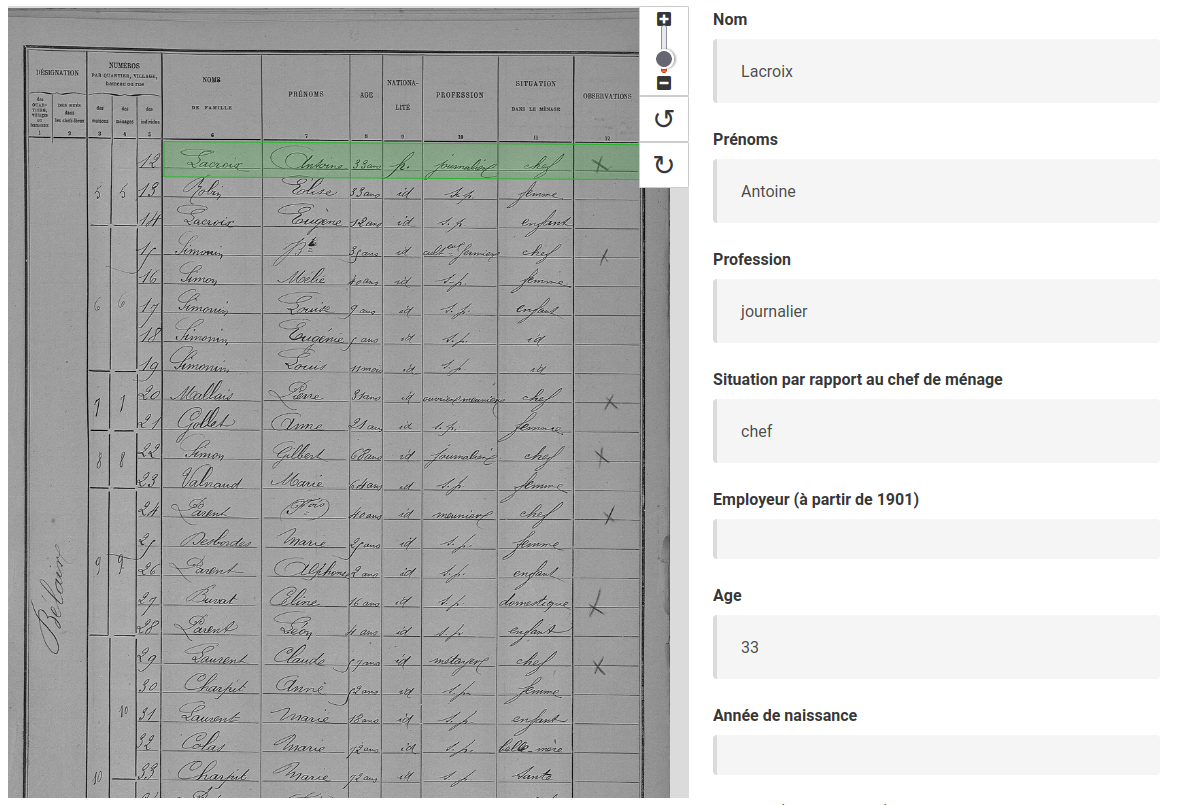}
         \caption{\textsc{Key-Value} mode to annotate the individuals’ information.}
         \label{fig:callico_individuals}
    \end{subfigure}%
    \hfill 
    \begin{subfigure}[b]{0.48\textwidth}
         \centering
         \includegraphics[width=\linewidth]{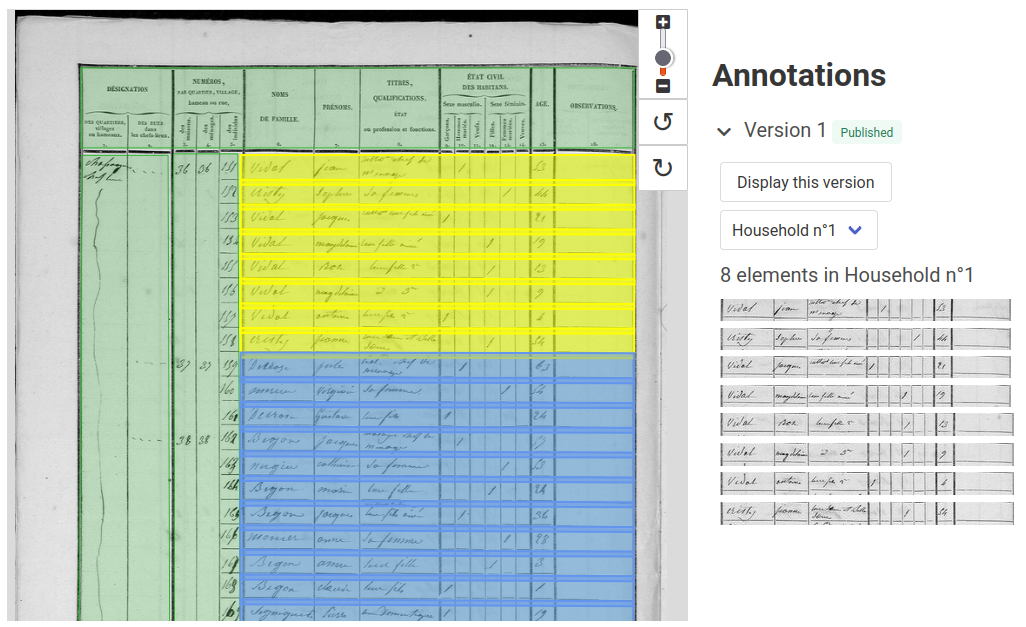}
         \caption{\textsc{Grouping} mode to group individuals into households.}
         \label{fig:callico_households}
    \end{subfigure}
    \caption{Callico interfaces for annotating information on individuals and grouping them into households.}
    \label{fig:callico}
\end{figure}

We selected 100 single pages from 11 pilot departmental archives for annotation. This selection was carefully chosen to include all years covered by the study and to accurately reflect the diversity of page appearance, image quality and table templates present in the archives. These images served as the basis for manual transcription tasks carried out on Callico \cite{callico2024}, an open-source document image annotation platform, using two specific annotation modes. First, the \textsc{Key-Value} mode was used to annotate the individual level information. In this mode, the annotator is presented with a full list page, with a highlighted zone corresponding to an individual entry, and is prompted to enter the relevant details into a designated form, as presented on Figure \ref{fig:callico_individuals}. Second, the \textsc{Grouping} mode was used to construct the household groupings present in the nominative lists. In this mode, the entire page is displayed with individual zones highlighted, as shown on Figure \ref{fig:callico_households}. Annotators are tasked with selecting zones that belong to the same household, in order to reconstruct of household units from the disjointed individual entries. In addition, we assigned a class to each selected page in order to train the page classification model described in Section \ref{sec-classification}.  \\

Throughout the project, 22 annotation campaigns were conducted - two for each of the 11 selected departmental archives - resulting in an annotated dataset for model training. The first type of campaign, which focused on detailed annotation of individual information, resulted in 33,815 rows of table data. For the household grouping efforts, a total of 532 pages were annotated. Importantly, the majority of these annotations underwent a moderation process where they were either validated or corrected by experts to ensure the highest possible accuracy and reliability of the ground-truth data. 


%
\section{Information extraction workflow}
\label{sec-ie-workflow}

This section describes the different models used to extract the personal data. To process only the list pages and extract the individual information, we start the processing by applying an image classification model, as described in Section \ref{sec-classification}. A text recognition model is then applied directly to the pages to extract the individual information. This model and its training parameters and performance are described in detail in Section \ref{sec-recognition}. 

\subsection{Page classification}
\label{sec-classification}

Since our study focuses on the pages of the nominal lists, and in order to save processing time, we only send images containing list pages to the recognizer. We therefore trained an image classification model with the following classes:
\begin{itemize}
    \item The \textsc{Front} class corresponds to the first page of a register, which contains all the information about the year, the commune, the department and also some instructions for filling in the nominative lists;
    \item The \textsc{List} class corresponds to pages of nominative lists with information on individuals, organized by household and street;
    \item The \textsc{Recap} pages contain various tables summarizing information about the population of the parish;
    \item The \textsc{Totals} pages contain the total number of houses, households, individuals, but also men and women in the commune;
    \item The \textsc{Other} class contains all other images such as blank pages, black pages or handwritten tables that do not correspond to nominative lists, summaries, or totals.
\end{itemize}

\subsubsection{Training configuration}

To train a model, we chose to fine-tune the classification model pre-trained on ImageNet \cite{deng2009} available in YOLOv8\footnote{\url{https://docs.ultralytics.com/tasks/classify/}}. We started from the \textsc{YOLOv8x-cls} model and fine-tuned on Socface images during 200 epochs with early stopping and a batch size of 4. The model was trained on square images of size $1024 \times 1024$ pixels. The data used is the same as that selected for Callico annotation to train the recognition model, to which we have added pages from classes other than \textsc{List}. In total, we have 1,285 pages, divided into 899 in the training set, 193 for the validation and 193 in the test set.


\subsubsection{Page classification results}

The performance results of the model were satisfactory, particularly in terms of minimizing classification ambiguities regarding the \textsc{List} class. The accurate identification of this class is critical, as these pages are subsequently processed by the information extraction model, which requires a high degree of precision and recall to ensure comprehensive data capture. As shown in Table \ref{tab:results-classification}, the model demonstrates exceptional efficiency, achieving precision and recall metrics of at least 99\% for the \textsc{List} class. 

\begin{table}[ht]
\caption{Results obtained by the image classification model.}
\begin{subtable}[t]{0.52\linewidth}
    \centering
    \caption{Results obtained by the image classification model for each set and class. The results on the training set are not shown because the model obtained 100\% for the precision, recall and F1-score for all classes.}
    \label{tab:results-classification}
    \setlength{\tabcolsep}{4pt}
    \begin{tabular}{l|rrrrrr}
        \toprule
        \multirow{2}{*}{\textbf{Class}} & \multicolumn{3}{c}{\textbf{Validation}} & \multicolumn{3}{c}{\textbf{Test}} \\
        & \textbf{P} & \textbf{R} & \textbf{F1} & \textbf{P} & \textbf{R} & \textbf{F1} \\
        \midrule
        \textsc{Front} & 1.0 & 1.0 & 1.0 & 0.93 & 1.0 & 0.97 \\
        \textsc{List} & 1.0 & 0.99 & 1.0 & 1.0 & 0.99 & 0.99 \\
        \textsc{Recap} & 0.92 & 1.0 & 0.96 & 0.91 & 0.83 & 0.87 \\
        \textsc{Totals} & 1.0 & 0.92 & 0.96 & 1.0 & 1.0 & 1.0 \\
        \textsc{Other} & 0.88 & 1.0 & 0.93 & 0.56 & 0.71 & 0.63 \\
        \bottomrule
    \end{tabular}
\end{subtable}%
\hfill
\begin{subtable}[t]{0.42\linewidth}
    \centering
    \caption{Confusion matrix of the test set.}
    \label{tab:classification-confusion-matrix}
    \setlength{\tabcolsep}{4pt}
    \renewcommand{\arraystretch}{1.5}
    \begin{tabular}{ccccccc}
        \multicolumn{1}{c}{} &\multicolumn{1}{c}{} &\multicolumn{5}{c}{Truth} \\
        & & \rotatebox{90}{\textsc{Front}} & \rotatebox{90}{\textsc{List}} & \rotatebox{90}{\textsc{Recap}} & \rotatebox{90}{\textsc{Totals }} & \rotatebox{90}{\textsc{Other}} \\
        \hhline{~~-----}
        \multirow{5}{*}{\rotatebox{90}{Predicted}} & \multicolumn{1}{c|}{\textsc{Front}} & \cellcolor[RGB]{228,239,249}14 & & & & \multicolumn{1}{c|}{} \\
        & \multicolumn{1}{c|}{\textsc{List}} & & \cellcolor[RGB]{8,48,107}\textcolor{white}{145} & & & \multicolumn{1}{c|}{\cellcolor[RGB]{245,249,254}2} \\
        & \multicolumn{1}{c|}{\textsc{Recap}} & & & \cellcolor[RGB]{234,242,251}10 & & \multicolumn{1}{c|}{\cellcolor[RGB]{245,249,254}2} \\
        & \multicolumn{1}{c|}{\textsc{Totals}} & & & & \cellcolor[RGB]{230,240,249}13 & \multicolumn{1}{c|}{} \\
        & \multicolumn{1}{c|}{\textsc{Other}} & \cellcolor[RGB]{246,250,255}1 & & \cellcolor[RGB]{246,250,255}1 & & \multicolumn{1}{c|}{\cellcolor[RGB]{241,247,253}5} \\
        \hhline{~~-----}
    \end{tabular}
    \renewcommand{\arraystretch}{1}
\end{subtable}
\end{table}


Furthermore, analysis of the confusion matrix shown in Table \ref{tab:classification-confusion-matrix} reveals that the model faces more challenges in classifying the \textsc{Other} class, which is characterized by its considerable diversity. This category combines data that includes tables that are often misidentified as summary pages, as well as pages with printed text that resemble front pages, leading to classification ambiguities and, consequently, reduced performance metrics within this specific class.

Notwithstanding the model's limitations in accurately classifying the \textsc{Other} class, its ability to identify list pages remains commendably high, making it sufficiently capable for the purposes of classifying and earmarking pages for subsequent processing by the information extraction model outlined in the following section.

\subsection{Handwritten table recognition}
\label{sec-recognition}

The information about the individuals is presented in a table where the individuals are grouped into households. 
Given the scale of the project and the diversity of the documents, it was not feasible to develop and maintain a processing chain comprising multiple models. We therefore chose the DAN model \cite{coquenet2023-2} to perform a full table recognition, which not only extracts the text from the table, but also tags the extracted text in order to categorize the predicted text at the same time \cite{tarride2023-1}. 

The advantage of this method is that it does not require any segmentation of the page nor the table, as it works directly on the whole page. In addition, some information is marked by vertical lines or ditto labels, so processing the whole page allows better interpretation of these labels compared to, for example, table row-level processing, where the model has much less context to interpret the content of a cell. A second advantage of this method is that there is no need to apply a second model later to label the information, as it is categorized directly at the same time as the text is recognized. 

Finally, this model can also be used to predict data in a structured way. In fact, by adding a token indicating the head of the household before the individual information, we are able to directly structure individuals into households without any other model. This structuring involves a post-processing step consisting of going through all the pages of the register, in the correct reading order, to reconstruct households spanning two pages. 

\subsubsection{Label generation}

\begin{figure}[ht]
    \centering
    \includegraphics[width=\linewidth]{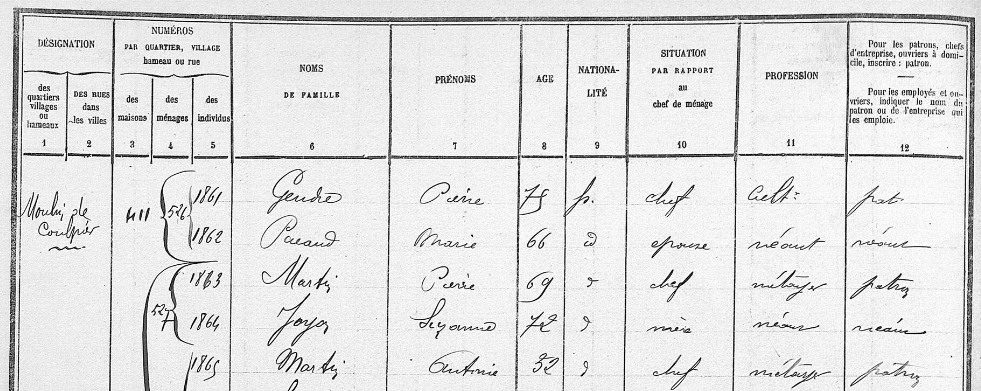}
    \caption{Table header and first rows of a table from the census of the commune of Neuilly-le-Réal (department of Allier) in 1901. The label used to train the model for this part of the table is: \\ \footnotesize \texttt{
    {\color{BlueGreen}<s-h>}Gendre {\color{DarkOrchid}<f>}Pierre {\color{OliveGreen}<o>}cultivateur {\color{Dandelion}<l>}chef {\color{WildStrawberry}<e>}patron {\color{LimeGreen}<a>}75 {\color{Maroon}<n>}française\\
    {\color{NavyBlue}<s>}Paraud {\color{DarkOrchid}<f>}Marie {\color{OliveGreen}<o>}néant {\color{Dandelion}<l>}épouse {\color{WildStrawberry}<e>}néant {\color{LimeGreen}<a>}66 {\color{Maroon}<n>}idem\\
    {\color{BlueGreen}<s-h>}Martin {\color{DarkOrchid}<f>}Pierre {\color{OliveGreen}<o>}métayer {\color{Dandelion}<l>}chef {\color{WildStrawberry}<e>}patron {\color{LimeGreen}<a>}69 {\color{Maroon}<n>}idem\\
    {\color{NavyBlue}<s>}Joyoz {\color{DarkOrchid}<f>}Suzanne {\color{OliveGreen}<o>}néant {\color{Dandelion}<l>}mère {\color{WildStrawberry}<e>}néant {\color{LimeGreen}<a>}72 {\color{Maroon}<n>}idem\\
    ...}\\
    \normalsize
    Note that the order of the entities in the labels is always the same and does not always correspond to the order in which the information appears in the images, as there are multiple templates.}
    \label{fig:label_example}
\end{figure}

To extract information from individuals and group them into households, we use a unique text recognition model. To train it, we constructed the ground truth transcriptions as described below and shown in Figure \ref{fig:label_example}:
\begin{itemize}
    \item Each piece of information annotated in the Callico form is preceded by a token indicating the type of information ({\footnotesize\texttt{{\color{BlueGreen}<s-h>}, {\color{NavyBlue}<s>}, {\color{DarkOrchid}<f>}, {\color{OliveGreen}<o>}, {\color{Dandelion}<l>}, {\color{WildStrawberry}<e>}, {\color{LimeGreen}<a>}, {\color{Maroon}<n>}}} in the Figure);
    \item The names of individuals listed as 'heads of household' are preceded by a token ({\footnotesize\texttt{{\color{BlueGreen}<s-h>}}}) that is different from the other members of the household ({\footnotesize\texttt{{\color{NavyBlue}<s>}}}), in order to indicate the start of the household;
    \item All information about an individual is concatenated into a single string so that it always follows the same order, even if it is different from the order in the table;
    \item The transcriptions for each individual are themselves concatenated to represent the whole page in a single string.
\end{itemize}

Empty cells and rows are not annotated and not present in the transcription.

\subsubsection{Model and training configuration}

DAN \cite{coquenet2023-2} is an open-source attention-based Transformer model for handwritten text recognition that can work directly on pages. The encoder is fully convolutional, while the decoder is a Transformer network. It is trained with the cross-entropy loss function. The last layer is a linear layer with a softmax activation function that computes probabilities associated with each vocabulary character. 
We trained a DAN model on the annotated single pages for 1,000 epochs with early stopping and a batch size of 4. The model was trained on a single GPU A100 with 80Gb. To reduce the memory required for training, the images were resized so that their height was equal to 1900 pixels. Data augmentation was applied during training and the maximum number of tokens to be predicted was set to 2,800 according to the training data.

\subsubsection{Full-page recognition results}

\begin{table}[ht]
\caption{Results obtained by the information extraction model.}
\begin{subtable}[t]{0.35\linewidth}
    \centering
    \caption{Character Error Rate and Word Error Rate obtained by the information extraction model (\%).}
    \label{tab:results-dan-htr}
    \setlength{\tabcolsep}{4pt}
    \begin{tabular}{l|rr}
        \toprule
        \textbf{Set} & \textbf{CER} & \textbf{WER} \\
        \midrule
        \textsc{train} & 8.94 & 17.18 \\
        \textsc{validation} & 14.30 & 26.22 \\
        \textsc{test} & 14.47 & 27.05 \\
        \bottomrule
    \end{tabular}
\end{subtable}%
\hfill
\begin{subtable}[t]{0.6\linewidth}
    \centering
    \caption{Evaluation of entity recognition on the test set.}
    \label{tab:results-dan-ner}
    \setlength{\tabcolsep}{4pt}
    \begin{tabular}{l|rrrr}
        \toprule
        \textbf{Tag} & \textbf{P} & \textbf{R} & \textbf{F1} & \textbf{Support} \\
        \midrule
        \textsc{age} & 0.87 & 0.87 & 0.87 & 1,700 \\
        \textsc{birth\_date} & 0.97 & 0.99 & 0.98 & 558 \\
        \textsc{civil\_status} & 0.95 & 0.93 & 0.94 & 1,153 \\
        \textsc{employer} & 0.74 & 0.76 & 0.75 & 237 \\
        \textsc{firstname} & 0.94 & 0.93 & 0.94 & 2,371 \\
        \textsc{link} & 0.85 & 0.89 & 0.87 & 1,838 \\
        \textsc{lob} & 0.74 & 0.76 & 0.75 & 788 \\
        \textsc{nationality} & 0.67 & 0.73 & 0.70 & 1,287 \\
        \textsc{observation} & 0.37 & 0.10 & 0.16 & 141 \\
        \textsc{occupation} & 0.83 & 0.80 & 0.81 & 1,496 \\
        \textsc{surname} & 0.86 & 0.82 & 0.84 & 1,835 \\
        \textsc{surname\_house.} & 0.72 & 0.80 & 0.76 & 519 \\
        \midrule
        \textbf{Total} & \textbf{0.85} & \textbf{0.85} & \textbf{0.85} & \textbf{13,923} \\
        \bottomrule
    \end{tabular}
\end{subtable}
\end{table}

The performance of the text recognition and household grouping model is shown in Tables \ref{tab:results-dan-htr} and \ref{tab:results-dan-ner}. The CER obtained on the validation and test sets are 14.30\% and 14.47\% respectively. These values, which may seem rather high, reflect the quality of all the categories of information to be extracted at the level of the whole page. As these metrics are strongly affected by a shift in recognition: an extra word, for example, shifts the entire predicted sequence, they are very difficult to interpret. For this reason, the performance of each entity is presented in Table \ref{tab:results-dan-ner}. From this table, the F1 scores for all the fields, except the "Observation" category, ranged from 70\% for nationality to 98\% for year of birth. These high scores show that the model is robust and generic enough to handle a large number of documents, image qualities and table templates.

From the table, we can also see that the information contained in the "Observation" columns is very poorly recognized, with an F1 score of 16\%. This can be explained by the fact that this category is very poorly represented during training: it appears only 388 times in the manual annotations of the training set, which means that the information is present in about 1\% of the table rows.

Finally, although it definitely plays a role, performance in the other categories does not seem to be directly correlated with the number of elements in the training set. Our hypothesis is that the difference in performance between the different categories can be explained by several other factors:
\begin{itemize}
    \item Some entities are easier to recognize because the possible values are very limited: this is particularly the case for the age and year of birth categories;
    \item Others entities are more difficult to recognize because they may contain ditto entries, and some annotators have rewritten the text in the corresponding cell rather than annotating it as a ditto.
\end{itemize}

In order to improve performance and make the results easier to interpret, further standardization of the annotations would be necessary, particularly to reduce the impact of this last factor.

\subsubsection{Household extraction}

Table \ref{tab:results-dan-ner} also shows the performance on the household grouping task, which consists of predicting a different category for the surnames. We can see that 76\% of the households were correctly grouped, which seems quite good considering the difficulty of the task. In fact, in some lists, the information is clearly annotated with brackets. But this is not always the case, and sometimes the information is not directly annotated but has to be inferred from the 'link' category, making the task much more complex.

%
\section{Distributed processing on HPC}
\label{sec-full-corpus-processing}

The Socface project leverages the capabilities of Arkindex, an open source document processing platform that offers a comprehensive suite of functionalities including document organization, visualization, processing, and export. However, we are faced with the monumental task of processing approximately 30 million images. To meet the demanding computational requirements of this huge dataset, we rely on public High-Performance Computing (HPC) resources. However, integration with the HPC infrastructure imposes specific constraints: the compute nodes are isolated from the Internet, requiring pre-staging of data on specialized local storage and orchestration of job submissions through dedicated scheduling systems such as SLURM.

To overcome these limitations, we have developed a three-step strategy to extend Arkindex to take advantage of HPC resources:
\begin{itemize}
    \item Data preparation and pre-processing: Recognizing the lack of Internet connectivity on HPC computing nodes, the first stage is performed on front-end CPU nodes that do have Internet access. This step involves downloading the required dataset images from the IIIF server, along with essential processing metadata such as image dimensions. These elements are then stored on local storage, making them available to the HPC computing nodes for subsequent processing stages.
    \item Image processing: With the data pre-positioned on local storage, processing shifts to the HPC's GPU nodes. This stage uses the computing power of the GPUs to efficiently analyze the images. The results of this processing stage are encapsulated in JSON files, providing a structured representation of the results that can be easily transferred and interpreted in subsequent steps.
    \item Integration and monitoring of results: The final phase moves back to CPU nodes with Internet access. This is where the JSON files containing the processed data are uploaded to the Arkindex database. This step not only secures the processed data within Arkindex, but also facilitates real-time task status updates. Such updates are critical for monitoring the progress and success of processing tasks, providing insight into operational status, and ensuring that any necessary adjustments or re-processing can be addressed in a timely manner.
\end{itemize}

To implement these various steps, Arkindex's internal distributed task system has been significantly enhanced by integrating the PySlurm library\footnote{\url{https://pyslurm.github.io/}}, which enables seamless communication with SLURM. This key development has effectively enabled Arkindex to take advantage of the immense computing resources available in HPC environments, significantly increasing its processing capacity to meet the demands of large-scale projects.

We conducted a processing time evaluation using our distributed processing framework enabled by High Performance Computing (HPC) to manage the extraction of information from a batch of 450,000 images, processed using a distributed architecture that integrated 14 parallel processes on CPU nodes for the initial and final stages of the workflow, and used NVIDIA V100 GPUs to execute the deep learning model responsible for table recognition and entity typing.

The breakdown of processing times for each stage of the workflow is as follows:
\begin{itemize}
    \item Preprocessing Phase: This initial phase, mainly focused on image download, was completed in an average time of 1.6 seconds per image. 
    \item Table Recognition and Entity Typing: The core processing task of recognizing full-page tables and typing entities within these tables using our deep learning model took an average of 12.5 seconds per image.
    \item Post-processing stage: The final phase, which included uploading the results to the database along with text position, line-level text recognition and entity tagging, took an average of 7.2 seconds per image.
\end{itemize}
The entire batch of 450,000 images was processed in less than 8 days, demonstrating the efficiency and scalability of our distributed processing approach using HPC resources. 

%
\section{Conclusion}
\label{sec-conclusion}

In this paper, we have presented a comprehensive workflow designed to automatically extract information from individual census tables spanning 20 censuses over a century, structured to closely follow the original format of the source documents. This methodology has already proven its effectiveness on thousands of images and will be scaled up to process millions more from numerous French departmental archives by the end of the project.

Our achievement lies in the development of a unified model capable of handling a wide variety of image types, table structures and handwriting styles. Using a transformer-based architecture, this model allows direct processing of entire tables without the need for prior segmentation, significantly minimizing the potential for errors commonly associated with multi-step processing approaches. Careful label generation ensures comprehensive information extraction across all table variants, covering both the content and the familial arrangements of the listed individuals.

However, the current method has limitations, most notably the inability to process complete registers on a page-by-page basis while retaining the context of previously processed pages. This shortcoming requires additional post-processing to reassemble household units that span multiple pages. Future enhancements will focus on overcoming this challenge by enabling sequential processing of entire registers with the aim of preserving contextual continuity. We also plan to extend the processing to include address recognition, thereby facilitating the reconstruction of household compositions within individual houses, streets, hamlets, and sectors. 

%
\section{Acknowledgments}
\label{sec-acknowledgments}

The Socface project is funded by the French National Research Agency (ANR) under the fund ANR-21-CE38-0013. This work was granted access to the HPC resources of IDRIS under the allocation 2022-AD011013446 made by GENCI and was partially funded by the ACADIIE project "Compréhension automatique des documents d’archives pour l’extraction d’informations individuelles"  supported by a grant overseen by the French National Research Agency (ANR) as part of the France Relance program.

%
%
\bibliographystyle{splncs04}
\bibliography{main}

\begin{thebibliography}{10}
\providecommand{\url}[1]{\texttt{#1}}
\providecommand{\urlprefix}{URL }
\providecommand{\doi}[1]{https://doi.org/#1}

\bibitem{oliveira2018}
Ares~Oliveira, S., Seguin, B., Kaplan, F.: {dhSegment: A Generic Deep-learning Approach for Document Segmentation}. In: {16th International Conference on Frontiers in Handwriting Recognition (ICFHR)}. pp. 7--12 (Aug 2018)

\bibitem{bernard2023}
Bernard, G., Wall, C., Boillet, M., Coustaty, M., Kermorvant, C., Doucet, A.: {Text Line Detection in Historical Index Tables: Evaluations on a New French PArish REcord Survey Dataset (PARES)}. In: {Leveraging Generative Intelligence in Digital Libraries: Towards Human-Machine Collaboration}. pp. 59--75. Springer Nature Singapore (Dec 2023). \doi{10.1007/978-981-99-8085-7\_6}

\bibitem{biswas2022}
Biswas, S., Banerjee, A., Llad{\'o}s, J., Pal, U.: {DocSegTr: An Instance-Level End-to-End Document Image Segmentation Transformer}. In: arXiv preprint arXiv:2201.11438 (2022)

\bibitem{boillet2020}
Boillet, M., Kermorvant, C., Paquet, T.: {Multiple Document Datasets Pre-training Improves Text Line Detection With Deep Neural Networks}. In: {25th International Conference on Pattern Recognition (ICPR)}. pp. 2134--2141 (Jan 2021)

\bibitem{constum2022}
Constum, T., Kempf, N., Paquet, T., Traounez, P., Chatelain, C., Bree, S., Merveille, F.: {Recognition and Information Extraction in Historical Handwritten Tables: Toward Understanding Early 20th Century Paris Census}. In: {15th International Workshop on Document Analysis Systems (DAS)}. p. 143–157 (May 2022). \doi{10.1007/978-3-031-06555-2\_10}

\bibitem{coquenet2023-2}
Coquenet, D., Chatelain, C., Paquet, T.: {DAN}: a segmentation-free document attention network for handwritten document recognition. In: {IEEE Transactions on Pattern Analysis and Machine Intelligence}. pp. 1--17. Institute of Electrical and Electronics Engineers ({IEEE}) (Jan 2023). \doi{10.1109/tpami.2023.3235826}

\bibitem{coquenet2023-1}
Coquenet, D., Chatelain, C., Paquet, T.: {End-to-end Handwritten Paragraph Text Recognition Using a Vertical Attention Network}. In: {IEEE Transactions on Pattern Analysis and Machine Intelligence}. pp. 508--524 (Jan 2023). \doi{10.1109/TPAMI.2022.3144899}

\bibitem{deng2009}
{Deng}, J., {Dong}, W., {Socher}, R., {Li}, L., {Kai Li}, {Li Fei-Fei}: {ImageNet: A Large-scale Hierarchical Image Database}. In: {IEEE Conference on Computer Vision and Pattern Recognition (ICPR)}. pp. 248--255 (Jun 2009). \doi{10.1109/CVPR.2009.5206848}

\bibitem{gruning2018}
Grüning, T., Leifert, G., Strauß, T., Labahn, R.: {A Two-Stage Method for Text Line Detection in Historical Documents}. In: {International Journal on Document Analysis and Recognition (IJDAR)}. pp. 285--302 (Sep 2019)

\bibitem{he2015}
He, K., Zhang, X., Ren, S., Sun, J.: {Deep Residual Learning for Image Recognition}. In: {IEEE Conference on Computer Vision and Pattern Recognition (CVPR)}. pp. 770--778 (Jun 2016)

\bibitem{callico2024}
Kermorvant, C., Bardou, E., Blanco, M., Abadie, B.: {Callico: a Versatile Open-Source Document Image Annotation Platform}. In: {Sumbitted to ICDAR2024} (2024)

\bibitem{motte2007}
Motte, C., Vouloir, M.C.: {Le site cassini.ehess.fr. Un instrument d’observation pour une analyse du peuplement}. Bulletin du Comité français de cartographie  \textbf{191},  68--84 (2007)

\bibitem{ren2015}
Ren, S., He, K., Girshick, R., Sun, J.: {Faster R-CNN: Towards Real-Time Object Detection with Region Proposal Networks}. In: {28th International Conference on Neural Information Processing Systems (NIPS)}. p. 91–99 (Jun 2015)

\bibitem{smock2022}
Smock, B., Pesala, R., Abraham, R.: {PubTables-1M: Towards Comprehensive Table Extraction From Unstructured Documents}. In: {IEEE/CVF Conference on Computer Vision and Pattern Recognition (CVPR)}. pp. 4634--4642 (Jun 2022)

\bibitem{tarride2023balsac}
Tarride, S., Maarand, M., Boillet, M., McGrath, J., Capel, E., V\'{e}zina, H., Kermorvant, C.: Large-scale genealogical information extraction from handwritten quebec parish records. Int. J. Doc. Anal. Recognit.  \textbf{26}(3),  255–272 (jan 2023). \doi{10.1007/s10032-023-00427-w}, \url{https://doi.org/10.1007/s10032-023-00427-w}

\bibitem{tarride2023-1}
Tarride, S., Boillet, M., Kermorvant, C.: {Key-Value Information Extraction from Full Handwritten Pages}. In: {Document Analysis and Recognition - ICDAR 2023}. pp. 185--204. Springer Nature Switzerland (Aug 2023). \doi{10.1007/978-3-031-41679-8\_11}

\bibitem{vaswani2017}
Vaswani, A., Shazeer, N., Parmar, N., Uszkoreit, J., Jones, L., Gomez, A.N., Kaiser, L.u., Polosukhin, I.: {Attention is All you Need}. In: {31st International Conference on Neural Information Processing Systems (NIPS)}. p. 6000–6010 (Dec 2017)

\end{thebibliography}

\end{document}